# ChatGPT and a New Academic Reality: AI-Written Research Papers and the Ethics of the Large Language Models in Scholarly Publishing

Brady D. Lund[1*], Ting Wang[2], Nishith Reddy Mannuru[1], Bing Nie[3], Somipam Shimray[4], Ziang Wang[5]

[1]University of North Texas, Department of Information Science, Denton, TX, USA

[2]Emporia State University, School of Library and Information Management, Emporia, KS, USA

[3]Zhejiang Tongji Vocational College of Science and Technology, Hangzhou, China

[4]Babasaheb Bhimrao Ambedkar University, Department of Library and Information Science, Lucknow, India

[5]Baker University, School of Education, Baldwin City, KS, USA

*Corresponding Author, brady.lund@unt.edu*

**Abstract**

This paper discusses OpenAI's ChatGPT, a generative pre-trained transformer, which uses natural language processing to fulfill text-based user requests (i.e., a "chatbot"). The history and principles behind ChatGPT and similar models are discussed. This technology is then discussed in relation to its potential impact on academia and scholarly research and publishing. ChatGPT is seen as a potential model for the automated preparation of essays and other types of scholarly manuscripts. Potential ethical issues that could arise with the emergence of large language models like GPT-3, the underlying technology behind ChatGPT, and its usage by academics and researchers, are discussed and situated within the context of broader advancements in artificial intelligence, machine learning, and natural language processing for research and scholarly publishing.

ChatGPT and related technologies have been identified as disruptive innovations with the potential to revolutionize academia and scholarly publishing (Haque et al., 2022). As a natural language processing tool developed by OpenAI, ChatGPT can automate the preparation of essays and other scholarly manuscripts. However, the ethical implications of this technology and its underlying GPT-3 technology have not yet been fully considered (Gonzalez-Padilla, 2022). This paper addresses the significant ethical issues that could arise with the use of GPT-3 and places these concerns in the context of broader advancements in artificial intelligence, machine learning, and natural language processing for research and scholarly publishing.

Our analysis reveals that the potential for bias in the training data and coding process of AI-driven language models such as GPT-3 poses a threat to the integrity of science. Additionally, ethical considerations include issues of copyright, citation practices, and the potential impact on the "Matthew Effect" in scholarly publishing. Ownership of the content generated by the model and the use of third-party content within the generated manuscripts are also major ethical concerns. While ChatGPT can assist editors and peer reviewers in completing repetitive or tedious tasks, there is a risk that it may not mitigate existing biases and that human judgment calls are still necessary. Furthermore, the use of AI-driven language models raises concerns about the reproducibility and transparency of research.

It is crucial to carefully consider all of these ethical considerations to ensure that these technologies are used responsibly and ethically in the context of academic research and publishing. Researchers, publishers, and the developers of AI-driven language models must collaborate to establish guidelines and protocols to ensure that the use of these technologies is ethical, transparent, and accountable. Failure to do so may undermine public trust in the scientific process and have far-reaching consequences for the future of research and innovation.

## A Brief Introduction to Underlying AI Concepts

Artificial intelligence, machine learning, and natural language processing are rapidly advancing fields that are having a significant impact on a wide range of industries and applications (Wamba et al., 2021). Chatbots, which are computer programs designed to mimic human conversation, are one example of the practical uses of these technologies (Adamopoulou & Moussiades, 2020a). The following paragraphs provide a brief overview of artificial intelligence, machine learning, and natural language processing –as well as the ways in which chatbots are utilizing these technologies to interact with users in a more human-like manner – in order to ensure that all readers have a basically familiarity with the technology and concepts that underpin large language models like the Generative Pretrained Transformer and ChatGPT.

Artificial intelligence (AI) is a multidisciplinary and interdisciplinary field that has grown tremendously since the introduction of manually operated computers in the 1950s (Haenlein & Kaplan, 2019). It has the potential to revolutionize various industries, and is defined as any theory, method, or approach that assists machines, particularly computers, in analyzing, simulating, exploiting, and exploring human thought processes and behaviors (Lu, 2019). AI involves the simulation of intelligent behavior in machines, with the goal of creating machines that can mimic human intellect in important ways such as language comprehension, reasoning, and problem-solving (Chowdhary, 2020). Self-learning is a key component of AI, allowing systems to acquire new information and improve their knowledge-based judgments and conclusions through experience and data (Mintz & Brodie, 2019). Machine learning and deep learning techniques have become the standard approach to advancing AI, with artificial neural networks (ANNs) allowing robots to learn and reason like humans and perform increasingly complex tasks (Lu, 2019). AI has already had a significant impact on various industries such as pharmaceutical, industrial, financial, medical, and managerial, and is expected to play a crucial role in helping businesses of all sizes stay competitive in the global economy (Makridakis, 2017). AI encompasses a range of specialized domains including machine learning (ML) and natural language processing (NLP).

The volume of data being generated from various sources, including people, devices, and computers, continues to grow at an exponential rate (Beath et al., 2012). As the amount of data becomes too large for individuals to make sense of and draw meaningful insights from, it becomes necessary to automate systems that can learn from the data to provide valuable insights. This is where machine learning (ML) comes in. ML is a key component of artificial intelligence (AI) and involves the development of computational theories for learning processes that allow machines to learn from experience without being explicitly programmed to do so (Chowdhary, 2020; Mahesh, 2020). It is at the intersection of computer science and statistics and is used to create programs that can automatically learn from data, acquire knowledge from experience, and continuously improve their learning behavior to make predictions based on new data (Jordan & Mitchell, 2015). ML is a useful technique in various areas of AI, including computer vision, voice recognition, and natural language processing (NLP). In this paper, we focus on ChatGPT, an AI technology that uses NLP to enable computers to engage in natural language conversations (Radford et al., 2018). ML is an essential part of ChatGPT, as it allows the system to learn from data and improve its language processing abilities over time, leading to more effective communication and interaction between humans and computers.

Artificial intelligence (AI) has become a part of everyday life with the advancement of processing power and the use of intelligent agents (Adamopoulou & Moussiades, 2020b). Intelligent agents are programs that can act independently and make decisions based on their observations of the environment, human input, and internal knowledge. Chatbots, also known as conversational artificial intelligence bots, are a type of intelligent agent that can respond to conversations through text or voice as if they were sentient beings, and they have gained popularity due to their usefulness in various applications such as customer service, healthcare, education, and personal support (Brandtzaeg & Følstad, 2017; Nagarhalli et al., 2020). Chatbots have been developed using natural language processing (NLP) techniques, which allow them to understand and interpret human language input (Khanna et al., 2015). In recent years, chatbots have become more popular, particularly among younger generations who prefer instant, one-on-one communication through short messages (Lokman & Ameedeen, 2018). This paper focuses on ChatGPT, a chatbot that uses NLP and AI to generate natural language conversations, and specifically on how it can be used in academia to create and write research and scholarly articles, and the ethical issues associated with this development.

# Introducing ChatGPT

OpenAI is a research laboratory that has made significant contributions to the field of artificial intelligence, including the development of the highly advanced language model, GPT-3. In addition to GPT-3, OpenAI has also released ChatGPT, a chatbot that uses natural language processing to generate responses to user inputs. Both GPT-3 and ChatGPT have garnered significant attention and have the potential to revolutionize a wide range of language-related tasks. In this section, we will delve into the details of OpenAI, GPT-3, and ChatGPT, exploring their capabilities, limitations, and potential applications.

**OpenAI and Early Developments**

Established in 2015, OpenAI is a research laboratory focused on the development of AI products for the common good (OpenAI, 2022). The laboratory has received significant support from individuals such as founding donor Elon Musk and the Microsoft Corporation, which invested one billion U.S. dollars in exchange for exclusive access to some of OpenAI's products (Brockman et al., 2016). As a result, the laboratory has made rapid progress in the development of its AI technologies. OpenAI has released a number of machine learning products for the general public, with DALL-E and ChatGPT being among the most well-known (Devlin et al., 2018).

DALL-E is a machine learning technology that generates images based on user inputs and was made widely available to the public, preceding ChatGPT in its availability (Marcus et al., 2022). It uses Artificial Neural Networks with multimodal neurons to understand and create novel images. These neurons are so tolerant of different expressions of a concept that they open up a world of variation (Goh et al., 2021). This innovation, introduced by OpenAI, represents how the lab seeks to change the game with smart AI. Availability to the public, key to the popularity of DALL-E, is also responsible for the rapid popularity of ChatGPT, which gained over one million unique users in less than one week after its launch (Mollman, 2022).

**ChatGPT**

Generative Pre-Trained Transformer, or GPT, is a language model that understands human inputs (as described with DALL-E above) and then produces response text that is nearly indistinguishable from natural human language (Dale, 2021). The concepts behind GPT are not too complex, by NLP/ML standards, thought they are refined by OpenAI (Radford et al., 2018). The creation and fine-tuning of the algorithm occurs in a couple of steps:

- Generative, unsupervised, pretraining, where the data used for training is unlabeled and learning occurs naturally (like when you walk into an entirely new situation with no background knowledge and fill in pieces as you explore) as opposed to supervised and guided training (where each aspect of the learning process is curated like a teacher in a classroom) (Erhan et al., 2010).

- Discriminative, supervised, fine-tuning, where, following the pretraining, the algorithm is refined by its creators to perform better on necessary tasks (Budzianowski & Vulić, 2019).

Figure 1 illustrates at a technical level how ChatGPT works, based on an initial GPT model, as well as the concepts of supervised fine-tuned modeling, rewards modeling, and proximal policy optimization modeling (PPO). As demonstrated here, development of a useful ChatGPT interface is an iterative process.

Figure 1. Diagram of the GPT Implementation Process

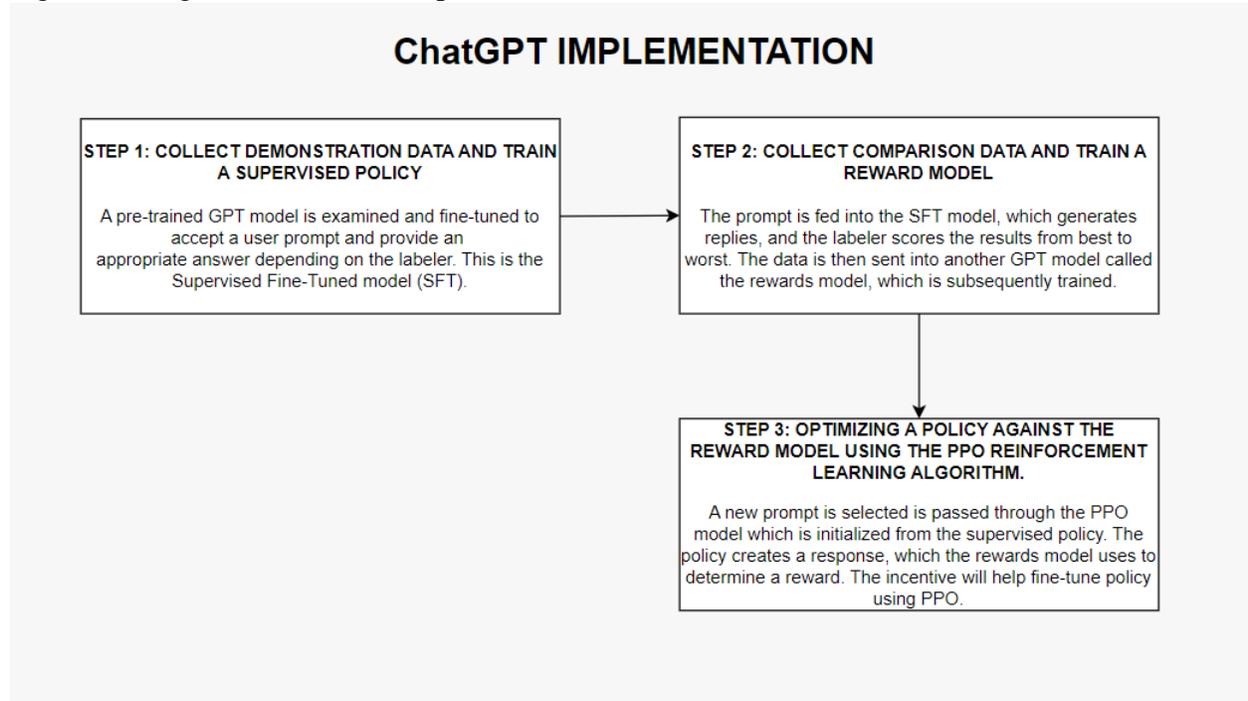

GPT stands out as a large language model due to its exceptional scale and the sheer amount of data used to train it. The algorithm underlying GPT has access to the entire Internet, which means that it is built upon billions of data sources, making it one of the largest language models in the world (Floridi & Chiriatti, 2020). GPT is a versatile tool that has been designed to perform various language-based tasks such as text generation, question answering, and translation. It uses deep learning techniques and sophisticated algorithms that enable it to understand the context of a piece of text and generate human-like responses, making it unique among language processing tools.

ChatGPT is a public tool developed by OpenAI that is based on the GPT technology (Kirmani, 2022). Essentially, it is a highly sophisticated chatbot that can fulfill almost any text-based request (Liu et al., 2021). However, ChatGPT is capable of much more than just answering simple questions. It can also fulfill more advanced requests, leveraging its extensive data stores and efficient design. For example, if you are unsure of what to write in a thank you card to a coworker, you can simply ask ChatGPT to write one for you, and it will quickly generate a well-written, multi-paragraph letter. ChatGPT can even help with more challenging tasks, such as composing a note to address a coworker's lack of productivity. It is only a small leap (and in the "mind" of ChatGPT, no leap at all) to ask GPT to write "an essay on the value of artificial intelligence."

In less than one minute, GPT can compose an essay of hundreds of words, written at professional researcher quality. An article could easily be written entirely by GPT by breaking the main topic into subtopics and then having GPT write each section. If the capacity of GPT is truly harnessed (through a full version that would allow for responses of several thousand words instead of just several hundred words), then an entire paper could be written in minutes with very minimal prompting from a researcher. This innovation could cut the time for composing research essays from several dozen hours to a couple dozen seconds, or even render the professional author/researcher obsolete.

Appendix 1 is an essay written by ChatGPT based on the simple prompt "write an essay about the value of AI." The essay is short, but is indistinguishable from a human-written response, perhaps even exceeding the average quality of a doctoral-level student.

Of course, there are a few limitations to GPT, even when ethical issues (which will be discussed in the rest of this paper) are set aside. Although the accuracy of natural language processing models is generally quite good, there are still errors that occur in interpreting meaning or creating accurate information – the models are not infallible, and GPT certainly has had its fair share of issues (Brown et al., 2020; Strubell et al., 2019). For instance, ambiguous terms, terms with multiple meanings (such as "construct"), as well as compound terms (such as "digital immigrant") can cause issues with interpretation of meaning by the GPT model. Additionally, these algorithms and data stores take a lot of energy to operate, especially at the scale that OpenAI is operating them (Zhou et al., 2021). OpenAI has also encountered several ethical pitfalls during the creation of GPT and the ChatGPT platform, which raise concerns about the organization's adherence to principles of responsible creation and maintenance of these technologies (Perrigo, 2023). There is also a risk that GPT could lead to biases a proliferation of misinformation, as many NLP algorithms are not yet skilled at understanding misinformation that may exist in its data stores and provide no resistance when asked to falsify or distort reality (Dale, 2017; Lucy & Bamman, 2021).

**Comparison of GPT/ChatGPT to Other Existing Language Models**

Compared to other language models like BERT, RoBERTa, and XLNet, ChatGPT and GPT-3 have several advantages. One of GPT-3's key strengths is its scale, with billions of parameters and access to vast amounts of data, it can perform an extensive range of language-related tasks with high accuracy. Additionally, GPT-3 is highly versatile and can be fine-tuned for various tasks such as translation, summarization, and question answering. It also uses data efficiently and achieves good results with relatively small amounts of training data. Furthermore, GPT-3 can generate human-like language, making it challenging to distinguish from text written by a human. While other language models may have similar objectives, none have achieved the same level of mastery as GPT-3 in these areas. Also, none that are close to achieving GPT-3's capabilities are as widely available to the public (Nature Machine Intelligence Editorial Board, 2020; Liu et al., 2019; Liu et al., 2022; Yang et al., 2019; Elkins & Chun, 2020).

**Benefits of ChatGPT for Scholarly Publishing**

Before discussing some of the ethical issues and concerns that emerge with using ChatGPT in scholarly publishing, it is important to stress the benefits that ChatGPT and similar large language models (BERT, XLNet) may bring to this field. As the building blocks of academic databases and the research environment, journals can contribute to the published articles' peer review and review ethics. ChatGPT can assist editors in completing repetitive or tedious tasks (e.g., correcting grammatical errors) and avoid making biased judgments about articles (Hosseini et al., 2022; Stokel-Walker, 2023). However, it is worth noting that if biased individuals train ChatGPT, it is unclear whether or how ChatGPT mitigates existing issues. Editors may still need judgment calls to reduce bias. On the other hand, peer reviewers sometimes fail to provide solutions in review reports due to a lack of motivation or capability (Horner & Lines, 2019; Waggoner, 2018). ChatGPT can improve the availability of review reports by providing solutions based on article contents. Since the collaborative review process (the process by which reviews from peer reviewers, editors, and other contributors are poled to provide a comprehensive set of recommendations for authors) is the primary mechanism for defining and helping to form a cognitive community (Woods et al., 2023), ChatGPT's positive role in the process may have a positive impact on the academic community, the research environment, and society as a whole (Thigpen & Funk, 2019).

ChatGPT can support the dissemination and diffusion of new research ideas through the creation of better metadata, indexing, and summaries of research findings (Lund & Wang, 2023). Assuming there is trust and usability among the public, this may support the diffusion of research findings through the translation of research into lay language that can be understood by members of the general public (Wang et al., 2022). ChatGPT can also be used as a sort of recommender system, helping with the identification of relevant research studies based on a user's query. This would be particularly helpful with highly interdisciplinary topics where searching across multiple indices and use of many synonymous search terms related to the jargon of different disciplines could be simplified by utilizing the capabilities of an interface like ChatGPT.

Certainly, ChatGPT can also benefit authors when used responsibly. ChatGPT should not be a substitute for one's knowledge of a topic, but it could be used to save time and expenses by composing descriptions of findings and structuring a paper according to various journal style guidelines. This is different from the use of ChatGPT for plagiarism – which will be discuss in the following sections – in that the platform is being used to save time and improve quality of communication, not reproduce unoriginal ideas (Gilat & Cole, 2023). An author would input paragraphs they have already written and then ask ChatGPT to "revise to improve clarity." Similarly, ChatGPT could be used for translation or revision of imperfect English in manuscripts in order to attain publication in higher quality journals (Jiao et al., 2023).

**Ethical Issues with Using ChatGPT for Scholarly Publishing**

The development of ChatGPT and other AI-driven technology brings with it great power and thus great responsibility. In light of this, ethical considerations must be taken into account. Although ChatGPT's massive Internet data set offers a wide range of perspectives and voices, Emily Bender, a computational linguist, has noted that size does not guarantee diversity (Bender et al., 2021). Additionally, AI-driven language models have been described as "stochastic parrots," regurgitating what they hear, often distorted by randomness (Hutson, 2021). Therefore, any ethical considerations of ChatGPT must also consider its potential for distortion. From an ethical standpoint, research papers created using ChatGPT can be seen as unoriginal and potentially problematic. Several studies have revealed that the training data and coding process of language models such as GPT-3, which are usually sourced from large web-based datasets, can contain bias with regard to gender, race, ethnicity, and disability status (Basta et al., 2019; Founta et al., 2018; Hutchinson et al., 2020; Tan & Celis, 2019; Zhao et al., 2019). Such a bias can be inadvertently perpetuated when these models are used to generate academic research, leading to the dissemination of hidden and unwitting prejudice. For thousands of years, human science has relied on verifiable evidence and an established system of checks and balances to ensure accuracy and fairness in the results of research. However, the increasing use of AI-generated research papers, which can be quickly produced in large numbers, poses a threat to the integrity of science by introducing potential biases and errors that are difficult to identify and correct (Muller, 2021). This can lead to further inequities in research outcomes and a potential undermining of the foundations of scientific knowledge.

In this section, the ethical implications of using ChatGPT and related technologies in academia and scholarly research and publishing will be examined. Specifically, issues of copyright, citation practices, and the potential impact on the "Matthew Effect" in scholarly publishing will be focused on. The potential impact of ChatGPT on research productivity and the value placed on human expertise will also be considered. It is important to carefully consider these ethical considerations in order to ensure that these technologies are used responsibly and ethically in the context of academic research and publishing.

**Authorship, Copyright, and Plagiarism**

Authorship attribution is a major issue concerning the generation of new knowledge by intelligent agents. There may be questions about the ownership of the content generated by the model (Schönberger, 2018). If a user provides input data to the model and the model generates content based on that input, it could be argued that the user owns the copyright to the generated content. However, if the model generates content independently of user input, the input provided is very limited (e.g., "write an essay on this topic"), or if the content is significantly edited by someone other than the user, it may be more difficult to determine ownership (Yanisky-Ravid, 2017). In these cases, it may depend upon the agreement reached with the developer of the model that generates the content; it may be necessary, at minimum, to include the model as a coauthor of the manuscript (Hugenholtz & Quintais, 2021). It is beneficial to contact the developer, should there be any question relating to the extent of the model's involvement in the creation of the knowledge.

In terms of copyright, there may be concerns about the use of third-party content within the generated manuscripts (Beaza-Yates, 2022). If the language model incorporates quotes, data, or other materials from external sources, it is important to ensure that this use is in compliance with copyright laws and that proper attribution is given (Gillotte, 2019). Depending on the nature of the use, it may be necessary to obtain permission from the copyright owner or to rely on a defense such as fair use. When using a tool like ChatGPT, it can be impossible to know the extent to which data from the training set may be quoted or otherwise used in the output generated by the tool (Dehouche, 2021). Some language models are trained with data from a specific source, like Wikipedia articles. In these cases, it would be relatively easy to identify the original source. However, models like GPT-3 are trained on a massive corpus of data from across the Internet, making the capacity to track the source virtually impossible. In other words, it may be possible that the model itself could be responsible for infringing on copyright and the risk from this infringement may be passed on to the author if they replicate it in a scholarly publication.

Plagiarism concerns arise from copyright issues, and AI has been previously known in journalism for plagiarism. As a result, the ethics of using ChatGPT to generate scientific articles have been widely discussed (Anderson et al., 2023). Plagiarism is not limited to copying text but also includes paraphrasing text, methods, graphics, ideas, and any other product of intelligence that belongs to another person (Gasparyan et al., 2017). However, ChatGPT-generated texts are based on published literature, and citation practice is encouraged in scholarly publishing. Therefore, if the original authors are credited in the text, plagiarism may not be involved (Pertile et al., 2015). Furthermore, ChatGPT's working principle allows it to access, understand, and synthesize the best human thoughts in just a few seconds (Dehouche, 2021).

Some publishers are considering removing open-access scientific research papers to prevent AI, such as ChatGPT, from accessing the articles, in an attempt to mitigate ethical concerns (Anderson et al., 2023). However, if publicly funded research and relevant papers are not available to the public, it may lead to different ethical discussions about open access policies (Dehouche, 2021). Other publishers have simply made clear their policies regarding the usage (or suspected usage) of ChatGPT and similar large language models in scholarly publications. For instance, editors from the renowned publication *Science* have prohibited the use of any text generated by ChatGPT or any other AI tools in papers published in the journal (Thorp, 2023).

**Citation Practices**

Citation practices in academia are an essential aspect of scholarly work, as they serve several purposes. Citing sources demonstrates the writer's expertise in their field, showing that they are familiar with the existing research on a particular topic (Santini, 2018). It also showcases the breadth and extent of the research that has been conducted in a particular field, as the writer is able to refer to multiple sources in their work (Hyland, 1999). Citing sources is also a way to respect the work of others, as it gives credit to the researchers whose ideas and findings are being used in the writer's own work (Ha, 2022).

Providing accurate and detailed citations is also important for maintaining the integrity of academic research. When readers or editors have doubts about the validity of a claim or argument, they can use the provided citations to verify the sources and assess the credibility of the work (Santini, 2018). This is particularly important in the context of academic progression and tenure, as citations are often used to evaluate, assess, and rank the work of researchers (Moher et al., 2018).

ChatGPT can utilize natural language processing (NLP) to scan academic papers and suggest appropriate citations for identified source materials (King, 2022). This tool has the potential to significantly streamline the citation process and reduce errors, as it can help researchers identify the sources they should be citing and suggest the proper formatting for the citations. Additionally, ChatGPT may also help researchers discover new sources of information and stay up to date with current advances in their fields.

However, it is worth noting that ChatGPT has been found to produce academic essays with missing references (see Appendix 1 as an example). While ChatGPT may be able to provide in-text references in future versions, it is important to consider the potential consequences of relying solely on an automated tool like ChatGPT instead of reviewing the literature in depth (King, 2023). The lack of references in early versions of ChatGPT could lead to disorder in scholarly publishing, as the integrity and credibility of the research may be called into question without proper citations.

Scholarly publishing platforms like ChatGPT have the potential to impact the "Matthew Effect," which refers to the tendency for successful researchers with high citation counts to continue to be successful and cited frequently, while lesser-known researchers struggle to gain recognition and citations (Merton, 1968). This phenomenon can perpetuate existing inequalities in academia (Perc, 2014). Google Scholar uses a ranking system based on citations, and the Matthew Effect is particularly noticeable, with highly cited articles appearing first in search results. As a result, most people only look at the first few pages of search results, leading to a self-reinforcing cycle in which articles with the most citations continue to receive more citations (Perc, 2014). Consider when one enters "COVID-19" in Google Scholar (or ChatGPT), they receive mostly outdated articles from 2020 and 2021, written by U.S. authors, but these articles have a lot of citations so are ranked highly; this, in turn, leads to these same outdated articles continuing to be cited as opposed to those which may be more relevant with recent developments with the disease.

Platforms like ChatGPT, which use citation counts as a factor in determining which publications to cite, may exacerbate this effect. Therefore, it is crucial that researchers continue to engage in careful and thorough review of the literature, even when using tools like ChatGPT to assist with the citation process. This will help to ensure the quality and rigor of academic work and prevent the perpetuation of inequalities in the field (Lund, 2022).

**Impact on Academic Job Expectations, Tenure, and Promotion**

Evaluation of a researcher's work for the purposes of hiring, promotion, and tenure often focuses on factors such as the number and value of funded grants, the number of published papers, and the number of citations (Moher et al., 2018). Miller et al. (2011) have noted that tenure-track faculty often feel pressure to publish a large number of papers in peer-reviewed journals, as those who do not publish in such journals may be denied tenure. This emphasis on publication has led to the enduring belief that "publish or perish" is an important principle in academia (Caplow & McGee, 1958). However, the focus on the number of publications and the prestige of the journals in which they are published can stifle creativity and innovation, as editors and reviewers may be more likely to favor work that supports mainstream theories and methods (Augier et al., 2005; Bedeian, 2004). Limitations on creativity and innovation in academic research can lead to a lack of meaning and relevance, which can hinder scientific progress (Bedeian, 1996). Additionally, the emphasis on publication may lead to a disconnect between academic research and practice, as researchers may prioritize their own standing among their peers rather than the promotion of best practices (Alutto, 2008; Miller et al., 2011).

ChatGPT is a machine learning tool that uses regression language modeling to predict future words based on a preexisting knowledge of language (Olsson & Engelbrektsson, 2022). In the context of academic writing, ChatGPT generates new papers based on existing papers and researchers' needs. This could potentially exacerbate the lack of innovation and disconnection from practice in academic research (Kaltenbrunner et al., 2022). While there has been longstanding debate about the importance of publications and citations in academia, the advent of ChatGPT may be a turning point in this debate.

There are several steps that could be taken to address the challenges posed by ChatGPT. First, academic journal publishers could work with computer science professionals to develop anti-ChatGPT software similar to adblockers, which would be able to detect papers generated by ChatGPT (Abd-Elaal et al., 2022). This could be achieved through the following process:

- Upload a paper to the software.
- The software will identify keywords in the paper.
- Use AI to generate relevant papers by following multiple paths, using the identified keywords as input.
- Compare the similarity between the uploaded paper and the generated papers to identify any potential match with ChatGPT-generated content.

Second, academic and research institutions and academic journals could encourage more creative and innovative research. This could help to add vitality to the academic community and broaden the range of research topics being explored. It could also reduce the likelihood that papers submitted to journals were produced by ChatGPT.

Third, changing the criteria for evaluating tenure or reevaluating the purpose of tenure in higher education and research institutions could be a fundamental solution to the ethical issues of scholarly publishing that arise from ChatGPT. Tenure is typically granted to protect academic freedom and allow faculty to explore and express controversial or unpopular ideas without fear of retribution (American Association of University Professors, n.d.; Nolan, 2004). However, if researchers are motivated primarily by the pursuit of tenure and are not concerned with the quality or relevance of their work, the purpose of academic research – to increase our understanding of the world and find ways to improve it – may be compromised (University of North Carolina, n.d.). Instead of relying solely on the number of papers published and the prestige of the journals in which they appear, higher education and research institutions could consider alternative criteria for evaluating tenure, such as the potential impact and relevance of the research to practice or its potential to make significant contributions to the field. By shifting the focus away from quantity and prestige and towards the quality and relevance of research, institutions can discourage the use of ChatGPT and encourage more ethical practices in scholarly publishing.

## Discussion

The comparison of GPT/ChatGPT to other existing language models has demonstrated the strengths of these models in various language-related tasks. GPT-3 has shown itself to be highly versatile, efficient in using data, and capable of generating human-like language, making it a valuable tool for tasks such as translation, summarization, and question answering (Liu et al., 2022). Meanwhile, ChatGPT has the potential to improve research productivity and the quality of academic publications by streamlining the citation process and helping researchers accurately identify and properly format citations. These capabilities make GPT/ChatGPT useful and valuable tools for researchers and academics.

However, the use of GPT/ChatGPT also raises several ethical concerns that must be considered (Zech, 2021). One issue is the ownership of the content generated by these models, as it is unclear who holds the rights to the generated text. There may also be concerns about the incorporation of third-party materials, such as quotes or data, and the need to ensure compliance with copyright laws and proper attribution. The use of GPT/ChatGPT may impact traditional practices and the evaluation of research, as it has the potential to streamline the citation process and may raise questions about the value placed on human expertise (Etzioni, 2017). While ChatGPT can be a useful tool for helping researchers identify and properly format citations, it is important to consider the potential for reliance on automated tools and the impact on traditional practices. It is crucial to address these ethical issues in order to ensure responsible and ethical use of GPT/ChatGPT in academia (Hancock et al., 2020).

The ethical concerns related to the use of GPT/ChatGPT, such as ownership of generated content and compliance with copyright laws, are applicable to AI, NLP, and chatbots as a whole (Jarrahi et al., 2022; Jobin, 2019). These concerns apply to any situation in which AI and chatbot technology is used to generate content or perform tasks that may have legal or ethical implications (Wang et al., 2022). In terms of ownership of generated content, the issue of who holds the rights to text or other content produced by AI and chatbots is a broader concern that applies to the use of these technologies in various contexts. Similarly, the need to ensure compliance with copyright laws and proper attribution is a concern that is relevant to any situation in which AI and chatbots are used to incorporate third-party materials (Hristov, 2016). The impact of AI and chatbots on traditional citation practices and the evaluation of research is also a concern that is applicable to these technologies as a whole (Cox, 2022). As AI and chatbots become more prevalent, it is important to consider the potential implications for traditional practices and to ensure that the value placed on human expertise is appropriately balanced with the use of these technologies.

There are also limitations to this paper, as the comparison of GPT/ChatGPT to other language models may only be applicable at the time of writing. As new language models are developed, the relative strengths and weaknesses of GPT/ChatGPT may change. Future research could explore the use of GPT/ChatGPT in conjunction with other language models or technologies in order to enhance their capabilities and performance. Additionally, it would be worthwhile to investigate the use of GPT/ChatGPT in different tasks and domains, as well as to consider the full range of existing language models. By expanding the scope of the study and exploring these areas, it may be possible to gain a more comprehensive understanding of the strengths and limitations of GPT/ChatGPT and other language models.

## Conclusion

ChatGPT and related technologies have the potential to significantly impact academia and scholarly research and publishing. However, it is important to carefully consider the ethical implications of these technologies, particularly in regard to the use of GPT-3 by academics and researchers. While ChatGPT and GPT-3 represent major advancements in artificial intelligence, machine learning, and natural language processing, it is necessary to ensure that they are used ethically and responsibly for scholarly research and publishing. Many questions about the ethics of using GPT in academia and its impact on research productivity remain unanswered. This paper aimed to provide a comprehensive overview of the current state of these discussions and to encourage further exploration of the ethical considerations surrounding the use of GPT and similar technologies in academia.

# References

Abd-Elaal, E., Gamage, S., Mills, J. E. (2022). Assisting academics to identify computer generated writing. *European Journal of Engineering Education, 47*(5), 725-745. https://doi.org/10.1080/03043797.2022.2046709

Adamopoulou, E., & Moussiades, L. (2020a). An overview of chatbot technology. In *IFIP International Conference on Artificial Intelligence Applications and Innovations* (pp. 373-383). Springer.

Adamopoulou, E., & Moussiades, L. (2020b). Chatbots: History, technology, and applications. *Machine Learning with Applications, 2*, article 100006. https://doi.org/10.1016/j.mlwa.2020.100006

Alutto, J.A. (2008). Final Report of the AACSB International Impact of Research Task Force. The Association to Advance Collegiate Schools of Business, Tampa, FL. https://www.aacsb.edu/insights/reports/impact-of-research

American Association of University Professors. (n.d.). Tenure. https://www.aaup.org/issues/tenure#:~:text=Why%20is%20tenure%20important%3F,conduct%20research%20in%20higher%20education.

Anderson, N., Belavy, D. L., Perle, S. M., Hendricks, S., Hespanhol, L., Verhagen, E., & Memon, A. R. (2023). AI did not write this manuscript, or did it? Can we trick the AI text detector into generated texts? The potential future of ChatGPT and AI in Sports & Exercise Medicine manuscript generation. BMJ Open Sport & Exercise Medicine, 9(1), e001568.

Augier, M., March, J. G., & Sullivan, B. N. (2005). Notes on the evolution of a research community: Organization studies in Anglophone North America, 1945–2000. *Organization Science, 16*(1), 85-95. https://doi.org/10.1287/orsc.1040.0108

Basta, C., Costa-jussà, M. R., & Casas, N. (2019). Evaluating the Underlying Gender Bias in Contextualized Word Embeddings. *Proceedings of the Workshop on Gender Bias in Natural Language Processing, 1*, 33–39. https://doi.org/10.18653/v1/W19-3805

Beath, C., Becerra-Fernandez, I., Ross, J., & Short, J. (2012). Finding value in the information explosion. *MIT Sloan Management Review, 53*(4), 18-20.

Bedeian, A. G. (1996). Thoughts on the making and remaking of the management discipline. *Journal of Management Inquiry, 5*(4), 311-318. https://doi.org/10.1177/105649269654003

Bedeian, A. G. (2004). Peer review and the social construction of knowledge in the management discipline. *Academy of Management Learning & Education, 3*(2), 198-216.

Bender, E. M., Gebru, T., McMillan-Major, A., & Shmitchell, S. (2021). On the Dangers of Stochastic Parrots: Can Language Models Be Too Big? *Proceedings of the ACM*

*Conference on Fairness, Accountability, and Transparency*, *2021*, 610–623. https://doi.org/10.1145/3442188.3445922

Brandtzaeg, P. B., & Følstad, A. (2017). *Why people use chatbots. In International conference on internet science* (pp. 377-392). Springer.

Brockman, G., Cheung, V., Pettersson, L., Schneider, J., Schulman, J., Tang, J., & Zaremba, W. (2016). Openai gym. arXiv. https://doi.org/10.48550/arXiv.1606.01540

Brown, T., Mann, B., Ryder, N., Subbiah, M., Kaplan, J. D., Dhariwal, P., ... & Amodei, D. (2020). Language models are few-shot learners. *Advances in Neural Information Processing Systems, 33*, 1877-1901.

Budzianowski, P., & Vulić, I. (2019). Hello, it's GPT-2--how can I help you? towards the use of pretrained language models for task-oriented dialogue systems. arXiv. https://doi.org/10.48550/arXiv.1907.05774

Caplow, T., & McGee, R.J. (1958). *The Academic Marketplace*. New York, NY: Basic Books.

Cherian, A., Peng, K. C., Lohit, S., Smith, K., & Tenenbaum, J. B. (2022). Are Deep Neural Networks SMARTer than Second Graders?. arXiv. https://doi.org/10.48550/arXiv.2212.09993

Chiusano, F. (2022, September 20). A Brief Timeline of NLP. NLPlanet. https://medium.com/nlplanet/a-brief-timeline-of-nlp-bc45b640f07d

Chowdhary, K. (2020). Natural language processing. In *Fundamentals of artificial intelligence* (p. 603-649). Springer.

Cox, A. (2022). How artificial intelligence might change academic library work: Applying the competencies literature and the theory of the professions. *Journal of the Association for Information Science and Technology*. https://doi.org/10.1002/asi.24635

Dale, R. (2017). NLP in a post-truth world. *Natural Language Engineering, 23*(2), 319-324. https://doi.org/10.1017/S1351324917000018

Dale, R. (2021). GPT-3 What's it good for? *Natural Language Engineering, 27*(1), 113-118. https://doi.org/10.1017/S1351324920000601

Dehouche, N. (2021). Plagiarism in the age of massive Generative Pre-Trained Transformers (GPT-3). *Ethics in Science and Environmental Politics, 21*, 17-23. https://doi.org/10.3354/esep00195

Devlin, J., Chang, M. W., Lee, K., & Toutanova, K. (2018). Bert: Pre-training of deep bidirectional transformers for language understanding. arXiv. https://doi.org/10.48550/arXiv.1810.04805

Elkins, K., & Chun, J. (2020). Can GPT-3 pass a Writer's turing test? *Journal of Cultural Analytics, 5(*2), 17212. https://doi.org/10.22148/001c.17212

Erhan, D., Bengio, Y., Courville, A., Manzagol, P., & Vincent, P. (2010). Why does unsupervised pre-training help deep learning. *Journal of Machine Learning Research, 11*, 625-660.

Etzioni, O. (2017). AI zooms in on highly influential citations. *Nature, 547*, article 32. https://doi.org/10.1038/547032a

Floridi, L., & Chiriatti, M. (2020). GPT-3: Its nature, scope, limits, and consequences. *Minds and Machines, 30*(4), 681-694. https://doi.org/10.1007/s11023-020-09548-1

Founta, A.-M., Djouvas, C., Chatzakou, D., Leontiadis, I., Blackburn, J., Stringhini, G., Vakali, A., Sirivianos, M., & Kourtellis, N. (2018). Large Scale Crowdsourcing and Characterization of Twitter Abusive Behavior (arXiv:1802.00393). arXiv. http://arxiv.org/abs/1802.00393

Gasparyan, A. Y., Nurmashev, B., Seksenbayev, B., Trukhachev, V. I., Kostyukova, E. I., & Kitas, G. D. (2017). Plagiarism in the context of education and evolving detection strategies. *Journal of Korean Medical Science, 32*(8), 1220-1227.

Gillotte, J. L. (2019). Copyright infringement in ai-generated artworks. *UC Davis Law Review, 53*, 2655-2676.

Gilot, R., & Cole, B. J. (2023). How will artificial intelligence affect scientific writing, reviewing and editing? The future is here. *Arthroscopy*. https://doi.org/10.1016/j.arthro.2023.01.014

Goh, G., Cammarata, N., Voss, C., Carter, S., Petrov, M., Schubert, L., Radford, A., & Olah, C. (2021). *Multimodal neurons in artificial neural networks*. Retrieved from https://doi.org/10.23915/distill.00030

González-Padilla, D. A. (2022). Concerns About the Potential Risks of Artificial Intelligence in Manuscript Writing. *The Journal of Urology*, 10-1097. https://doi.org/10.1097/JU.0000000000003131

Ha, T. (2022). An explainable artificial-intelligence-based approach to investigating factors that influence the citation of papers. *Technological Forecasting and Social Change, 184*, article 121974.

Haenlein, M., & Kaplan, A. (2019). A brief history of artificial intelligence: On the past, present, and future of artificial intelligence. *California Management Review, 61*(4), 5-14. https://doi.org/10.1177/0008125619864925

Hancock, J. T., Naaman, M., & Levy, K. (2020). AI-mediated communication: Definition, research agenda, and ethical considerations. *Journal of Computer-Mediated Communication, 25*(1), 89-100. https://doi.org/10.1093/jcmc/zmz022

Haque, M. U., Dharmadasa, I., Sworna, Z. T., Rajapakse, R. N., & Ahmad, H. (2022). " I think this is the most disruptive technology": Exploring Sentiments of ChatGPT Early Adopters using Twitter Data. arXiv. https://doi.org/10.48550/arXiv.2212.05856

Hirschberg, J., & Manning, C. D. (2015). Advances in natural language processing. Science, 349(6245), 261–266. https://doi.org/10.1126/science.aaa8685

Horner, R. D., & Lines, L. M. (2019). Anatomy of Constructive Peer Review. *Medical Care, 57*(6), 399-400.

Hristov, K. (2016). Artificial intelligence and the copyright dilemma. *IDEA: The Intellectual Property Law Review, 57*(3), 431-454.

Hugenholtz, P. B., & Quintais, J. P. (2021). Copyright and artificial creation: Does EU copyright law protect AI-assisted output? *International Review of Intellectual Property and Competition Law, 52*, 1190-1216. https://doi.org/10.1007/s40319-021-01115-0

Hutchinson, B., Prabhakaran, V., Denton, E., Webster, K., Zhong, Y., & Denuyl, S. (2020). Social Biases in NLP Models as Barriers for Persons with Disabilities. *Proceedings of the Annual Meeting of the Association for Computational Linguistics, 58*, 5491–5501. https://doi.org/10.18653/v1/2020.acl-main.487

Hutson, M. (2021). Robo-writers: The rise and risks of language-generating AI. *Nature, 591*(7848), 22–25. https://doi.org/10.1038/d41586-021-00530-0

Hyland, K. (1999). Academic attribution: Citation and the construction of disciplinary knowledge. *Applied Linguistics, 20*(3), 341-367.

Jarrahi, M. H., Lutz, C., Boyd, K., Oesterlund, C., & Willis, M. (2022). Artificial intelligence in the work context. *Journal of the Association for Information Science and Technology*. https://doi.org/10.1002/asi.24730

Jiao, W., Wang, W., Huang, J., Wang, X., & Tu, Z. (2023). Is ChatGPT a good translator? A preliminary study. *Arxiv*. https://doi.org/10.48550/arXiv.2301.08745

Jobin, A., Ienca, M., & Vayena, E. (2019). The global landscape of AI ethics guidelines. *Nature Machine Intelligence, 1*, 389-399. https://doi.org/10.1038/s42256-019-0088-2

Jordan, M. I., & Mitchell, T. M. (2015). Machine learning: Trends, perspectives, and prospects. *Science, 349*(6245), 255–260. https://doi.org/10.1126/science.aaa8415

Kaltenbrunner, W., Pinfield, S., Waltman, L., Woods, H. B., & Brumberg, J. (2022). Innovating peer review, reconfiguring scholarly communication: An analytical overview of ongoing peer review innovation activities. *Journal of Documentation, 78*(7), 429-449. https://doi.org/10.1108/JD-01-2022-0022

Kalyanathaya, K. P., Akila, D., & Rajesh, P. (2019). Advances in natural language processing: A survey of current research trends, development tools and industry applications. *International Journal of Recent Technology and Engineering, 7*(5C), 199-201.

Khanna, A., Pandey, B., Vashishta, K., Kalia, K., Pradeepkumar, B., & Das, T. (2015). A Study of Today's A.I. through Chatbots and Rediscovery of Machine Intelligence. International Journal of U- and e-Service, Science and Technology, 8(7), 277–284. https://doi.org/10.14257/ijunesst.2015.8.7.28

King, M. R. (2022). The future of AI in medicine: A perspective from a chatbot. *Annals of Biomedical Engineering*. https://doi.org/10.1007/s10439-022-03121-w

King, M. R. (2023). A conversation on artificial intelligence, chatbots, and plagiarism in higher education. *Cellular and Molecular Bioengineering*. https://doi.org/10.1007/s12195-022-00754-8

Kirmani, A. R. (2022). Artificial intelligence-enabled science poetry. *ACS Energy Letters, 8*, 574-576. https://doi.org/10.1021/acsenergylett.2c02758

Liu, J., Shen, D., Zhang, Y., Dolan, W. B., Carin, L., & Chen, W. (2022). What Makes Good In-Context Examples for GPT-3?. *Proceedings of Deep Learning Inside Out (DeeLIO 2022): The 3rd Workshop on Knowledge Extraction and Integration for Deep Learning Architectures, 3*, 100-114.

Liu, X., Zheng, Y., Du, Z., Ding, M., Qian, Y., Yang, Z., & Tang, J. (2021). GPT understands, too. arXiv. https://doi.org/10.48550/arXiv.2103.10385

Liu, Y., Mittal, A., Yang, D., & Bruckman, A. (2022). Will AI console me when I lose my pet? Understanding perceptions of AI-mediated email writing. *Proceedings of the CHI Conference on Human Factors in Computing Systems, 2022*, article 474. https://doi.org/10.1145/3491102.3517731

Liu, Y., Ott, M., Goyal, N., Du, J., Joshi, M., Chen, D., ... & Stoyanov, V. (2019). Roberta: A robustly optimized bert pretraining approach. arXiv. https://doi.org/10.48550/arXiv.1907.11692

Lokman, A. S., & Ameedeen, M. A. (2018, November). Modern chatbot systems: A technical review. In *Proceedings of the future technologies conference* (pp. 1012-1023). Springer, Cham.

Lu, Y. (2019). Artificial intelligence: A survey on evolution, models, applications and future trends. *Journal of Management Analytics, 6*(1), 1–29. https://doi.org/10.1080/23270012.2019.1570365

Lucy, L., & Bamman, D. (2021). Gender and representation bias in GPT-3 generated stories. *Proceedings of the Workshop on Narrative Understanding, 3*, 48-55.

Lund, B. D. (2022). Is academic research and publishing still leaving developing countries behind? *Accountability in Research, 29*(4), 224-231. https://doi.org/10.1080/08989621.2021.1913124

Lund, B. D., & Wang, T. (2023). Chatting about ChatGPT: How may AI and GPT impact academia and libraries? *Library Hi Tech News*. https://doi.org/10.1108/LHTN-01-2023-0009

Mahesh, B. (2020). Machine learning algorithms-a review. *International Journal of Science and Research (IJSR), 9*, 381-386.

Makridakis, S. (2017). The forthcoming artificial intelligence (AI) revolution: Its impact on society and firms. *Futures, 90*, 46-60. https://doi.org/10.1016/j.futures.2017.03.006

Marcus, G., Davis, E., & Aaronson, S. (2022). *A very preliminary analysis of DALL-E 2*. ArXiv pre-print. Retrieved from https://doi.org/10.48550/arXiv.2204.13807

Merton, R. K. (1968). The Matthew Effect in Science: The reward and communication systems of science are considered. *Science, 159*(3810), 56-63.

Miller, A. N., Taylor, S. G., & Bedeian, A. G. (2011). Publish or perish: Academic life as management faculty live it. *Career Development International, 16*(5), 422-445.

Mintz, Y., & Brodie, R. (2019). Introduction to artificial intelligence in medicine. Minimally Invasive Therapy & Allied Technologies, 28(2), 73–81. https://doi.org/10.1080/13645706.2019.1575882

Mollman, S. (2022). *ChatGPT gained 1 million users in under a week*. Retrieved from https://www.yahoo.com/lifestyle/chatgpt-gained-1-million-followers

Müller, V. C. (2021). Ethics of Artificial Intelligence and Robotics. E. N. Zalta, The Stanford Encyclopedia of Philosophy (Summer 2021). Metaphysics Research Lab, Stanford University. https://plato.stanford.edu/archives/sum2021/entries/ethics-ai/

Nadkarni, P. M., Ohno-Machado, L., & Chapman, W. W. (2011). Natural language processing: An introduction. *Journal of the American Medical Informatics Association, 18*(5), 544–551. https://doi.org/10.1136/amiajnl-2011-000464

Nagarhalli, T. P., Vaze, V., & Rana, N. K. (2020). A Review of Current Trends in the Development of Chatbot Systems. *In 2020 International Conference on Advanced Computing and Communication Systems (ICACCS), 6,,* 706–710. https://doi.org/10.1109/ICACCS48705.2020.9074420

Nature Machine Intelligence Editorial Board. (2020). Next chapter in artificial writing. *Nature Machine Intelligence, 2*, 419.

Nolan, C. W. (2004). Tenure-Track or Tenure Trap?. In P.K. Shontz (Ed.), *The librarian's career guidebook* (p. 281-290). Lanham, MD: Scarecrow Press.

Olsson, A., & Engelbrektsson, O. (2022). *A thesis that writes itself: On the threat of AI-generated essays within academia*. (Bachelor Thesis). https://www.diva-portal.org/smash/get/diva2:1669744/FULLTEXT02

OpenAI. (2022). *OpenAI about page.* Retrieved from https://openai.com/about/

Perc, M. (2014). The Matthew effect in empirical data. *Journal of the Royal Society Interface, 11*(98). https://doi.org/10.1098/rsif.2014.0378

Perrigo, B. (2023). *OpenAI used Kenyan workers on less than $2 per hour to make ChatGPT less toxic*. Retrieved from https://time.com/6247678/openai-chatgpt-kenya-workers/

Pertile, S., Moreira, V. P., & Rosso, P. (2015). Comparing and combing content- and citation-based approaches for plagiarism detection. *Journal of the Association for Information Science and Technology, 67*(10), 2511-2526.

Radford, A., Narasimhan, K., Salimans, T., & Sutskever, I. (2018). *Improving language understanding by generative pre-training*. Retrieved from https://www.cs.ubc.ca/~amuham01/LING530/papers/radford2018improving.pdf

Santini, A. (2018). The importance of referencing. *The Journal of Critical Care Medicine, 4*(1), 3. https://doi.org/10.2478/jccm-2018-0002

Schönberger, D. (2018). *Deep copyright: Up- and downstream questions related to artificial intelligence (AI) and machine learning (ML).* SSRN. Retrieved from https://papers.ssrn.com/sol3/papers.cfm?abstract_id=3098315

Stokel-Walker, C., & Van Noorden, R. (2023). What ChatGPT and generative AI mean for science. *Nature, 614*(7947), 214-216.

Strubell, E., Ganesh, A., & McCallum, A. (2019). Energy and policy considerations for deep learning in NLP. *Proceedings of the Annual Meeting of the Association for Computational Linguistics, 57*, 3645-3650.

Tan, Y. C., & Celis, L. E. (2019). Assessing social and intersectional biases in contextualized word representations. Proceedings of the 33rd International Conference on Neural Information Processing Systems (pp.13230–13241). Curran Associates Inc.

Thigpen, C., & Funk, C. (2019). Most Americans says science has brought benefits to society and expect more to come. https://www.pewresearch.org/fact-tank/2019/08/27/most-americans-say-science-has-brought-benefits-to-society-and-expect-more-to-come/

Thorp, H. H. (2023). ChatGPT is fun, but not an author. *Science*, 379(6630), 313. https://doi.org/10.1126/science.adg7879

University of North Carolina. (n.d.). What is Sociology?. https://sociology.unc.edu/undergraduate-program/sociology-major/what-is-sociology/

Waggoner, A. (2018). Improving the quality of constructive peer feedback. *College Teaching, 66*(1), 22-23.

Wamba, S. F., Bawack, R. E., Guthrie, C., Queiroz, M. M., & Carillo, K. D. (2021). Are we preparing for a good AI society? A bibliometric review and research agenda. *Technological Forecasting and Social Change, 164*, article 120482. https://doi.org/10.1016/j.techfore.2020.120482

Wang, X., Lin, X., & Shao, B. (2022). Artificial intelligence changes the way we work: A close look at innovating with chatbots. *Journal of the Association for Information Science and Technology, 74*(3), 339-353. https://doi.org/10.1002/asi.24621

Woods, H. B., Brumberg, J., Kaltenbrunner, W., Pinfield, S., & Waltman, L. (2023). An overview of innovations in the external peer review of journal manuscripts. *Wellcome Open Research, 7*(82), 82.

Yang, Z., Dai, Z., Yang, Y., Carbonell, J., Salakhutdinov, R., & Le, Q. V. (2019). XLNet: generalized autoregressive pretraining for language understanding. *Proceedings of the 33rd International Conference on Neural Information Processing Systems, 33*, 5753-5763.

Yanisky-Ravid, S. (2017). *Generating Rembrandt: Artificial intelligence, copyright, and accountability in the 3A Era: The human-like authors are already here*. Michigan State Law Review, Review 659.

Zech, H. (2021). Liability for AI: Public policy considerations. *ERA Forum, 22*, 147-158. https://doi.org/10.1007/s12027-020-00648-0

Zhao, J., Wang, T., Yatskar, M., Cotterell, R., Ordonez, V., & Chang, K.-W. (2019). Gender Bias in Contextualized Word Embeddings. Proceedings of the 2019 Conference of the North American Chapter of the Association for Computational Linguistics: Human Language Technologies, Volume 1 (Long and Short Papers), 629–634. https://doi.org/10.18653/v1/N19-1064

Zhou, X., Chen, Z., Jin, X., & Wang, W. Y. (2021). HULK: An energy efficiency benchmark platform for responsible natural language processing. *Proceedings of the Conference of the European Chapter of the Association for Computational Linguistics: System Demonstrations, 16*, 329-336.

**Appendix 1. ChaptGPT's essay on the value of AI (written entirely by ChatGPT, without any human editing)**

Artificial intelligence (AI) refers to the development of computer systems that are able to perform tasks that normally require human intelligence, such as learning, problem-solving, and decision making. In recent years, AI has made significant strides in a wide range of fields, and its potential to revolutionize and improve many aspects of our lives is vast.

One of the main values of AI is its ability to automate and streamline processes, freeing up time and resources for more complex or high-value tasks. For example, AI can be used to analyze large amounts of data, identify patterns and trends, and make predictions or recommendations, all tasks that would be time-consuming and resource-intensive for humans to perform. This can enable organizations and individuals to make more informed decisions and to operate more efficiently.

Another key value of AI is its ability to augment and enhance human capabilities. For example, AI-powered assistants and chatbots can help people to access information and services more quickly and easily, while AI-powered tools and devices can help people to communicate, collaborate, and work more effectively.

AI also has the potential to drive innovation and progress in a wide range of fields, from healthcare and education to transportation and energy. For example, AI-powered tools and algorithms can help researchers and scientists to analyze data, make discoveries, and develop new therapies and treatments, while AI-powered systems and devices can help to improve the efficiency, safety, and sustainability of various industries and sectors.

In conclusion, AI has the potential to bring many benefits and improvements to our lives and society as a whole. By harnessing the power of AI and using it responsibly and ethically, we can help to shape a brighter and more prosperous future.